# A Reinforcement Learning Based Approach for Automated Lane Change Maneuvers

Pin Wang[1*], Ching-Yao Chan[1], Arnaud de La Fortelle[1,2]

*Abstract* — Lane change is a crucial vehicle maneuver which needs coordination with surrounding vehicles. Automated lane changing functions built on rule-based models may perform well under pre-defined operating conditions, but they may be prone to failure when unexpected situations are encountered. In our study, we proposed a Reinforcement Learning based approach to train the vehicle agent to learn an automated lane change behavior such that it can intelligently make a lane change under diverse and even unforeseen scenarios. Particularly, we treated both state space and action space as continuous, and designed a Q-function approximator that has a closed-form greedy policy, which contributes to the computation efficiency of our deep Q-learning algorithm. Extensive simulations are conducted for training the algorithm, and the results illustrate that the Reinforcement Learning based vehicle agent is capable of learning a smooth and efficient driving policy for lane change maneuvers.

## I. INTRODUCTION

Interests in autonomous vehicles have seen great increase in recent years, from automakers to high-tech companies and research institutions. While fully autonomous vehicles at Automation Level 5 per SAE-J3016 [1] may only be widely available in the more distant future, partially or highly automated features at Automation Level 2, 3, or 4 look promising to be commercially available in the near future. Experience from the mass deployment of Advanced Driver Assistance Systems (ADAS) and from extensive tests of autonomous driving show that "steady state" driving is operational and research is now focusing onto "transition" maneuvers that are intrinsically riskier. Lane changing is one fundamental and crucial function expected to be embedded in either an ADAS or a fully automated vehicle. The lane changing maneuver can be a demanding task because the vehicle needs to alertly watch the leading vehicle on its ego lane and surrounding vehicles on the target lane, and to perform proper actions according to the potential adversarial or cooperative reactions demonstrated by those relevant vehicles. A study shows that around 10 percent of all freeway crashes are caused by lane change maneuvers [2]. Automated vehicles should be equipped to address such challenging maneuvers.

1. Pin Wang, Ching-Yao Chan and Arnaud de La Fortelle are with California PATH, University of California, Berkeley, 1357 South 46th Street., Richmond, CA, 94804, US. {pin_wang, cychan, arnaud.delafortelle}@berkeley.edu.

2. A. de La Fortelle is also with MINES ParisTech, PSL, Research University, Centre for Robotics, 60 Bd. St. Michel, 75006, France.

Research on automated lane changing maneuvers has been extensively conducted and the work can broadly be divided into two functional categories: a decision-making module and a control execution module [3]. A decision-making module can be viewed as a strategic or tactical level function which issues a lane change command based on a planned route (e.g. exit the highway from an off-ramp forward) or a desired driving condition (e.g. drive at high speed by passing a slow vehicle in front). When a lane change command is given, the ego vehicle (i.e. lane changing vehicle) performs operational control to coordinate the longitudinal and lateral movements for a safe, smooth and efficient lane change maneuver.

A considerable body of literature is existent on the topic of the decision-making function, and it generally adopts data-driven methods that are typically based on the abundance of available training datasets. This topic deserves in-depth discussion on how the decision-making functionality can be enhanced by considering demands and constraints from strategic, tactic and even operational levels. However, to limit the scope of this paper, we will focus on the aspect of operational control, i.e. how the vehicle can automatically perform the lane change maneuver once it receives the command that comes from a decision-making module.

With regard to the lane changing maneuver, it is a typical time sequential problem where the completion of the task involves a sequence of actions, and the performance of the current action has an impact on the ultimate goal of the task (e.g. a successful lane change). Such kind of problems are quite suitable to be solved by machine learning techniques, particularly by Reinforcement Learning (RL). The focus of the paper is to demonstrate the application of Reinforcement Learning on finding an optimal driving policy (e.g. smooth, safe and efficient) for automated lane change maneuvers.

The remainder of the paper is organized as follows. A literature review of related work is given in Section II. Section III introduces the methodology and details of our algorithms. Simulation results are given in Section IV. Concluding remarks and discussions of future research avenues are provided in the last section.

## II. RELATED WORK

Traditional approaches for addressing the autonomous lane changing problem primarily depend on pre-defined rules and explicitly designed models. Most of these approaches introduce a virtual lane change trajectory or a series of way points for the ego vehicle to follow when a lane change process is initiated. In [4], a virtual reference lane change trajectory was established with a polynomial equation. The virtual curvature of the intended path, along with steering angle, was fed into a bicycle model to estimate the vehicle lateral positions. In [5], a number of way points with information acquired from Differential Global Positioning System and Real-time Kinematic devices were used to generate a guidance path when a vehicle embarked on a lane change. A common limitation in these approaches is the lack of flexibility in the planned trajectories under dynamic situations and diverse driving styles. Moreover, though it might work relatively well in predefined situations or within the model limits, it is far from adequate in handling situations that are out of the defined scope. This is also a clear limitation in the current time-receding optimization methods such as Model Predicted Control (MPC) as the optimization criteria may be too complex to be explicitly formulated for all scenarios, and such methods always involves the predictions of future trajectories. An alternative way is to connect all vehicles via cooperative techniques such as in [6], but in the present paper we deal with a stand-alone system.

Machine learning (ML) algorithms have the capability of dealing with unforeseen situations after being properly trained on a large set of sample data without explicitly specific design and programming rules beforehand. Vallon et al. [7] proposed to use Support Vector Machine to make the lane change decision tailored to the driver's individual driving preferences. After the lane change demand is generated, the maneuver is executed using a MPC. Bi et al. [8] moved further to apply machine learning algorithms on both the decision-making process and the lane changing execution process based on traffic simulation. Specifically, they trained a randomized forest model for decision-making and a neural network for prediction of vehicle velocities at the next step, not only for the ego vehicle but also for the follower vehicle, under an assumption of connected driving environment.

Another study combined the conventional control, a PID controller, with neural networks, for designing a self-tuning and adaptive lateral controller [9]. The inputs to the neural network are optimal lateral acceleration, real lateral acceleration and the error between these two values. This approach should theoretically work well, but may be difficult to implement due to the difficulty in obtaining optimal acceleration values in real-world driving situations.

Reinforcement learning, one promising category in the machine learning family, has the capability of dealing with time-sequential problems and seeking optimal policies for long-term objectives by learning from trials and errors, without resorting to an off-line collected database. RL has been extensively applied in robotics, e.g. [10], and video games, e.g. [11]. Recently, it has been used in automated vehicle field [12, 13]. However, the driving scenarios in these aforementioned studies were relatively simple because few interactions with surrounding vehicles were considered.

In one previous study [14], we used RL to train the vehicle agent to learn an optimal ramp-merge driving policy under interactive driving environment. We treated both the state space and action space as continuous to more realistically represent the real-world driving situations. The results demonstrated that with the reinforcement learning approach, the automated merge actions can take place safely, smoothly and promptly under an interactive driving environment. In that work, however, we did not take into account the lateral control issue as we assumed that the agent vehicle would follow the centerline of its travel lane. In this work, we expand the methodology we used in our previous work and adopt the RL architecture for the lane-change case with extended definitions in state space, action space and reward function. The design of our proposed framework could also work if more complex states and actions are defined, indicating promising perspectives.

III. METHODOLOGY

In a typical RL problem, the goal is to find an optimal policy $\pi^*$ which maps the states ($s \in S$) of the environment into actions ($a \in A$) that the agent takes at the corresponding time step in a way of maximizing a total expected return $G$ that is a cumulative sum of immediate rewards $r$ received over the completion of a task [15]. $S$ and $A$ here are state space and action space, respectively, and can be discrete or continuous in a specific problem. The reward $r(s,a)$ reflects the effects of an action $a$ in a given state $s$, whereas $G(s,a)$ is a long-term reward starting from a state $s$, taking an action $a$ and thereafter following a policy $\pi$.

In our case, the driving environment involves the interaction with other vehicles whose behaviors may be cooperative or adversarial. For example, when a vehicle reveals its intention of a lane change by turning on the turning signal, the lag vehicle on the target lane may cooperatively decelerate or change its path to yield, or it may adversarially accelerate just to deter the vehicle from cutting into its course of motion. Consequently, it is not trivial to model the environment explicitly with all possible future situations. Thus, we resort to a model-free approach to find the optimal policy.

Another point to adopt a RL based approach is that a traditional MPC based controller often uses explicitly defined sensor inputs, while the outcome from an image perception module is usually an extremely large feature map and might be fuzzy. MPC has difficulty in handling such a large and fuzzy set of inputs. In contrast, the RL agent can take in hundreds or even millions of features as inputs which do not necessarily have an explicit representation. In this sense, the RL approach can conveniently be connected with the perception module. Besides, from another perspective of a joint approach, a RL/ML module can be taken as a mediator component between a perception module and a traditional MPC module, as to take the perception outcome as input and output a reference guidance for the MPC controller.

### 3.1 Lane Change Controllers

It is a good practice to optimize the longitudinal and lateral control in one unified model as in a MPC controller, but it is also common to separate the complicated vehicle maneuvers into two correlated modules, i.e. a longitudinal control module and a lateral control module [4]. To explore the learning ability of a RL vehicle agent and demonstrate how the functionalities can be partitioned and addressed separately, we design two correlated controllers in our current work. Investigation of one integral controller is also worth pursuing and we defer it to our future work.

Since certain longitudinal control models are off-the-shelf and ready for use in application, we choose to leverage a well-developed car-following model, Intelligent Driver Model (IDM), to build the longitudinal controller. The lateral control is to be learned by RL with the consideration that most of those previously proposed lateral control models are far too theoretical or empirical to be applied on an autonomous vehicle.

There is also a gap-selection module working in parallell with the two controllers. After the vehicle get a lane change command, the gap selection module will check the availability of a safety gap formed by a leader vehicle and a follower vehicle on the target lane based on all the information of surrounding vehicles (e.g. speeds, accelerations, positions, etc.). If the gap is adequate enough to accommodate the speed difference under allowable maximum acceleration/deceleration and to ensure a minimum safety distance under current speed, it is considered as an acceptable gap, and then the lane change controllers will be initiated. Longitudinal and lateral controllers will interact with each other and function jointly to perform the overall lane change task while they are designed separately as individual modules.

### 3.2 IDM based Longitudinal Controller

IDM is a time-continuous car-following model for the simulation of highway and urban traffic. It describes the dynamics of relevant vehicles. Detailed information can be found in [16].

The IDM holds realistic properties compared with other intelligent models such as Gipps' model [17]. However, if it is applied directly, a phenomenon will appear where vehicles may travel at a relatively slow speed under moderate traffic conditions due to the safety constraints inherited in the equation, thereby, we modify the IDM to suit our purpose. As it is not a primary focus of this paper and the content space is limited, the details on the formulation and the illustrations of vehicle dynamics in car-following scenarios are omitted here. It is still worth mentioning that with the modified car-following model, the vehicle can perform reasonable longitudinal behaviors either in free traffic conditions or in interactive driving situations.

It is also worth noting that when the ego-vehicle is making a lane change, it may see two leading vehicles in the ego lane and the target lane during the transition. The IDM car-following model that we implement will allow the ego-vehicle to adjust its longitudinal acceleration by balancing between its two leaders, if observed, on its ego lane and the target lane. The smaller value will be used to weaken the potential discontinuity in vehicle acceleration incurred from lane change initiation. At the same time, the gap selection module is still working as a safety guard during the whole lane changing process to check whether the gap is still acceptable at each time step. If not, the decision-making module will issue a command to alter or abort the maneuver, and the control execution module will reposition in the original lane. In this way, the longitudinal controller takes the surrounding driving environment into account to ensure longitudinal safety, whereas the lateral controller directs the vehicle to intelligently merge into any accepted gap.

### 3.3 RL based Lateral Controller

In this section, we define the state space $S$, the action space $A$, the immediate reward function $r(s, a)$, and a model-free approach, Q-learning, to find the optimal driving policy.

#### 3.3.1 Action Space

In some RL studies, the action space is usually treated as discrete to make a problem more easily solvable, however, it might weaken the feasibility of the solution when applied to real-world problems. In our study, the lateral control in a lane changing process is of crucial importance since a slightly fallacious shift in steering may result in the vehicle drifting out of the lane or a significant disturbance of surrounding vehicles. Bearing this in mind, we design the lateral controller with a continuous action space to allow a realistic and smooth transition from one lane to another.

To make sure that the steering angle input is continuous and smooth, there should be no abrupt change in yaw rate or, in other words, the yaw acceleration does not fluctuate erratically. Thereby, we design the RL agent to learn the yaw acceleration, i.e., the action space is defined with vehicle yaw acceleration $a_{yaw}$.

$$a = a_{yaw} \in A \qquad (1)$$

#### 3.3.2 State Space

As mentioned earlier, the ego vehicle's interaction with surrounding vehicles is taken into account by the gap selection module and longitudinal controller, the lateral controller thereby only considers factors that affect the ego vehicle's lateral motion. For an integral controller to be developed in the future, all the relevant state information from all adjacent vehicles will be incorporated into the state space.

A successful lane change not only relates to vehicle dynamics, but also depends on road geometry, i.e., whether a lane change is performed on a straight segment or a curve. In our study, we define the state space with both vehicle dynamics and road information. To be specific, the state space includes the ego vehicle's speed $v$, longitudinal acceleration $a$, position $(x, y)$, yaw angle $\theta$, target lane $id$, lane width $w$, and road curvature $c$.

$$s = (v, a, x, y, \theta, id, w, c) \in S \qquad (2)$$

In this problem formulation, we suppose the required state information is available from sensor fusion modules which take in real-time data from on-board sensors, such as GPS, IMU, radar, camera, CAN, etc., and that it meets the desired accuracy requirements for the control purpose.

When the input state is in high dimensional space (e.g. from a vision perception module) or with measurement noise, the defined state space here can be adaptively expanded to a large size without changing the algorithm structure, which is one superiority of the RL approach.

*3.3.3 Reward Function*

Typically, in a lane change process our attention is on the safety, smoothness and efficiency of the maneuver. As the longitudinal module and the gap selection module have taken into account the safety concerns, smoothness and efficiency are considered by the lateral controller through the reward function. The components in the reward function are selected based on the most relevant variables to the action performance, and their weights are decided by trying different sets of parameters.

To be specific, the smoothness is evaluated by yaw acceleration $r_{acce}$ due to the consideration that yaw acceleration directly affects the magnitude of a shifting in the lateral movement.

$$r_{acce} = w_{acce} * f_{acce}(a_{yaw}) \qquad (3)$$

where $r_{acce}$ represents the immediate reward obtained from yaw acceleration, $w_{acce}$ is a weight which can be designed as a constant value or a function relevant with lateral position, and $f_{acce}$ is a function for evaluating $a_{yaw}$. We currently use a simple format as $r_{acce} = -1.0 * |a_{yaw}|$.

Another indicator of smoothness is the yaw rate $\omega_{yaw}$, which reflects the comfort of the driver in a lane change process since a higher yaw rate will result in a significant centrifugal pull force in driving. The function is given in (4).

$$r_{rate} = w_{rate} * f_{rate}(\omega_{yaw}) \qquad (4)$$

where $r_{rate}$ reflects the immediate reward obtained from yaw rate, $w_{rate}$ is a weight and can also be designed as a constant or a function, and $f_{rate}$ is a function for evaluating $\omega_{yaw}$. In our study, we apply $r_{rate} = -1.0 * |\omega_{yaw}|$.

The efficiency is assessed by the lane changing time consumed to complete the maneuver. Such a component is taken into account by adding a third term to the reward function, so as to avoid an overly long and extended lane change actions.

$$r_{time} = w_{time} * f_{lat}(\Delta d_{lat}) \qquad (5)$$

where $r_{time}$ is the immediate reward for evaluate the action efficiency, and $w_{time}$ is the weight, and $f_{lat}$ is a function of the current lateral deviation $\Delta d_{lat}$ to the target lane. The larger the $\Delta d_{lat}$ is, the longer the time is spent on lane changing. We use $r_{time} = -0.05 * |\Delta d_{lat}|$ for the efficiency assessment.

The immediate reward $r$ in a single step is a summation of the three parts. To assess the overall performance, we also need to calculate the total reward $R$ that is a cumulative return of immediate rewards over a complete lane changing process. Equally, the total reward can also be viewed as a composition of the three aforementioned individual parts: the total reward from yaw acceleration $R_{acce}$, the total reward from yaw rate $R_{rate}$, and the total reward from lane changing time $R_{time}$, as shown in (6).

$$R = \sum_{i=1}^{N}(r_{acce})_i + \sum_{i=1}^{N}(r_{rate})_i + \sum_{i=1}^{N}(r_{time})_i \qquad (6)$$

In the formulation, we define the rewards with negative values, also called "cost". The idea is that a cost can be considered as a penalty of an action if we want to evaluate the adverse impact of an action, such as unsmooth or inefficient of a lane change maneuver. In this way, the agent should be able to learn to avoid taking actions that result in a large penalty.

*3.3.4 Q-learning*

Q-learning is a model-free reinforcement learning technique which is used to find an optimal action selection policy through a Q-function by estimating the value of the total return, without waiting until the end of an episode to cumulate all the rewards.

Standard Q-learning algorithms, which try to learn the entire mapping from discrete state-action pairs to Q-values, will lose viability with the increased size of state-action spaces. An alternative solution is to use a Q-function approximator, normally modeled as neural networks, to output the Q values. This makes it possible to apply Q-learning to large table size problems, even when the state space is continuous. However, it cannot directly handle problems with continuous action space.

There are policy gradient based algorithms, e.g. actor-critic, to directly learn the policy without resorting to a value function [18,19]. Though such approaches can commendably avoid relying on a value function for policy improvement, it needs much effort on designing the policy network (as well as the value function network if in actor-critic algorithm). In this work, we put effort on modifying the Q-function network structure, to accommodate Q-learning to the continuous action space.

In particular, we design a Q-function which is quadratic in action so that the greedy action has a closed-form solution. The theoretical explanation is similar with the work in [20]. This variant of Q-learning avoids invoking a policy neural network and simplifies the learning algorithm. Mathematically, the Q-function approximator is expressed as follows

$$Q(s,a) = A(s) * (B(s) - a)^2 + C(s) \qquad (7)$$

where $A$, $B$, and $C$ are coefficients and designed with neural networks with the state information as input, as illustrated in Fig. 1. Because any smooth Q-function can be Taylor expanded in this form near the greedy action, there is not much loss in generality if we stay close to the greedy action in the Q-learning exploration.

In our model, $A$ is designed with a two-layer neural network with eight neurons in the input layer (i.e. the dimension of the state space $S$) and 100 neurons in the hidden layer. Particularly, $A$ is bounded to be negative with the use of a softplus activation function, multiplied by a negative sign, on the output layer. $C$ is also a two-layer neural network with the same number of neurons and layers as $A$, but it also takes in a terminal state as an indicator of the completion of a lane change. $B$ is in a heavily engineered form with three two-layer neural networks, which have the same structure of 8 neurons in the first layer and 150 neurons in the hidden layer, linked together to output a final yaw acceleration which is adaptable to different driving situations. One neural network is used to calculate preliminary yaw acceleration $a_{yaw\_pre}$. Another neural network is to calculate a factor called sensitivity factor $\beta_{sen}$ to compensate for the fix time interval in two consecutive steps. And the third neural network is to output a variable maximum yaw acceleration $a_{yaw\_max}$ that is used to give an adjustable boundary of the learned yaw acceleration according to the current input state. The abstract mathematical formulation of $B$ is in (8), and the structure of $B$ is illustrated in Fig. 2.

$$a_{yaw} = max\,(a_{yaw\_pre} * \beta_{sen}, a_{yaw\_max})$$

With the above design, the optimal action for a given state can be obtained from $B$. In the meanwhile, Q-values are calculated with additional information from $A$ and $C$. There is also a target Q-network, with the same structure but different parameters, to calculate target Q values. Weights are updated based on a loss defined between target Q-values and online Q-values with a RL learning technique called experience replay [21].

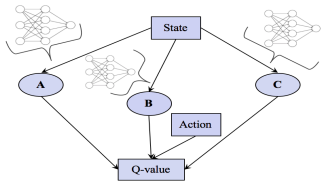

Figure 1. Structure of the Q-function approximator

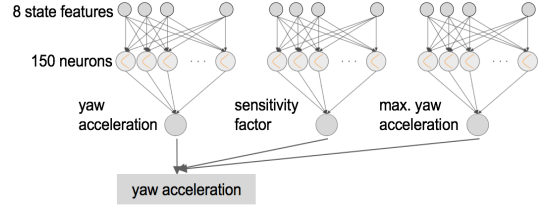

Figure 2. Structure design of $B$

## IV. SIMULATION AND RESULT

We test our proposed algorithms through a simulation platform where a learning agent is able to, on one hand, interact with the driving environment, and on the other hand improve itself by trials and errors from experiments. The simulation driving environment is a highway segment with three lanes on one direction. The segment length is 1000m and each lane width is 3.75m. Traffic on the highway can be customized as needed. For example, the initial speed, the departure time, and the speed limit of an individual vehicle can all be set to a random value within a reasonable and practical range to perform various longitudinal behaviors. In our current study, the departure interval is between 5s-10s, and individual speed limits are in a range of 80km/h-120km/h. All vehicles can perform practical car-following behaviors with the proposed IDM. Diverse traffic conditions can be generated with different sets of parameters. A scene of the simulation scenario is shown in Fig. 3.

In the training process, we set the training steps as 40,000, the time step interval $dt$ as 0.1s, the learning rate $\alpha$ as 0.01, and the discount factor $\gamma$ as 0.9. In total around 5000 vehicles executed the lane change maneuver in the training process. The loss and rewards gathered during training are shown in Fig. 4.

From the top graph in Fig. 4, we can observe that the loss curve shows an obvious downtrend indicating the convergence along with training steps. In the reward graphs, the total reward curve (upper left) shows the cumulative reward returns are increased during training, implying that the agent learned to maneuver with higher reward actions. The other three reward graphs demonstrate similar increasing trends as supplementary evidence to the learnability of expected behaviors.

Such training results give preliminary conclusion that the RL based vehicle agent is capable of learning the lane change policy with regard to our designed reward function under the proposed Q-function approximation architecture. More evaluation on the kinematic driving performance will be conducted in the future.

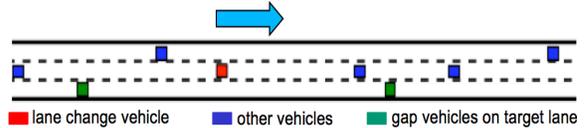

Figure 3. Simulation scenario

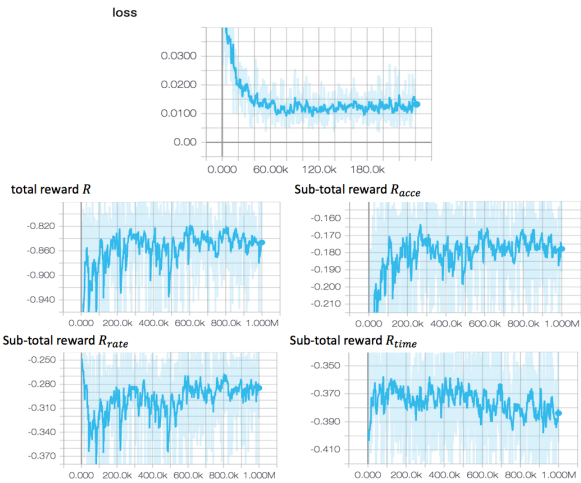

Figure 4. RL training loss and Q-values

## V. CONCLUSION AND DISCUSSION

In this work, we applied the Reinforcement Learning approach for learning the automated lane change behavior under interactive driving environment. The state space and action space are both treated as continuous to learn a more practical driving maneuver. A unique format of a quadratic function is used as the Q-function approximator, in which the coefficients are learned from neural networks. The reward function is defined with yaw rate, yaw acceleration and lane changing time for training a smooth and efficient lane change behavior.

To implement the proposed approach, we developed a simulation platform where diverse simulation scenarios can be generated by adjusting traffic density, initial speeds, speed limit, etc. Preliminary training results showed convergence of the loss and rewards as defined in the learning framework, indicating a promising attempt.

The next step for our research is to extend the testing of the RL agent under different road geometries and variant traffic flow conditions, in order to enhance its robustness and adaptability in complicated driving scenarios, which is a most advantageous feature of this method to deal with. Secondly, the lane changing performance (e.g. lateral position deviation and steering angle, etc.) along with the convergence trend will be compared with existing optimization-based approaches such as MPC, for validation and verification. In addition, another promising enhancement is to combine RL and MPC to make the best of both approaches. The proposed architecture can be established by extending the first layer of the neural networks to accommodate a large and fuzzy input state from an image module, and then output a reference guidance for a traditional optimization-based controller to issue a quick and reliable control command to the vehicle.

REFERENCE

1. Taxonomy and Definitions for Terms Related to Driving Automation Systems for On-Road Motor Vehicles: http://standards.sae.org/j3016_201609/.
2. S. Hetrick. Examination of driver lane change behavior and the potential effectiveness of warning onset rules for lane change or "side" crash avoidance systems. Dissertation, Virginia Polytechnic Institute & State University, 1997.
3. P. Hidas. Modeling vehicle interactions in microscopic simulation of merging and weaving. Transportation Research Part C: Emerging Technologies, Vol. 13, No. 1, 37–62, 2005.
4. M.L. Ho, P.T. Chan, A.B. Rad. Lane Change Algorithm for Autonomous Vehicles via Virtual Curvature Method. Journal of Advanced Transportation, Vol. 43, No. 1, pp. 47-70, 2007.
5. Y. Choi, K. Lim, J. Kim. Lane Change and Path Planning of Autonomous Vehicles using GIS. 12th International Conference on Ubiquitous Robots and Ambient Intelligence (URAI), Korea, 2015.
6. X. Qian, A. De La Fortelle, F. Moutarde. A hierarchical model predictive control framework for on-road formation control of autonomous vehicles. In Intelligent Vehicles Symposium (IV), 2016 IEEE, pages 376–381.
7. C. Vallon, Z. Ercan, A. Carvalho, F. Borrelli. A machine learning approach for personalized autonomous lane change initiation and control. IEEE Intelligent Vehicles Symposium (IV), Los Angeles, 2017.
8. H. Bi, T. Mao, Z. Wang, Z. Deng. A Data-driven Model for Lane-changing in Traffic Simulation. Eurographics/ ACM SIGGRAPH Symposium on Computer Animation, 2016.
9. G. Zhenhai. and W. Bing. An Adaptive PID Controller with Neural Network Self-Tuning for Vehicle Lane Keeping System. SAE Technical Paper, 2009-01-1482, 2009.
10. J. Schulman, P. Moritz, S. Levine, M. I. Jordan, P. Abbeel. High-Dimensional Continuous Control Using Generalized Advantage Estimation. 2015.
11. J. Schulman, S. Levine, P. Moritz, M. I. Jordan, P. Abbeel. (2015) Trust Region Policy Optimization. arXiv:1502.05477v5 [cs.LG].
12. A. Yu, R. Palefsky-Smith., R. Bedi. Deep reinforcement learning for simulated autonomous vehicle control. Stanford University. StuDocu. 2016.
13. A. Sallab, M. Abdou, E. Perot, S. Yogamani. End-to-End Deep Reinforcement Learning for Lane Keeping Assist. 30th Conference on Neural Information Processing Systems (NIPS), Barcelona, Spain, 2016.
14. P. Wang, C. Chan, Formulation of Deep Reinforcement Learning Architecture Toward Autonomous Driving for On-Ramp Merge. IEEE 20th International Conference on ITS, JAPAN, 2017.
15. R. S. Sutton, A. G. Barto. Book: Reinforcement Learning: An Introduction. 2016.
16. M. Treiber, A. Hennecke, D. Helbing. Congested traffic states in empirical observations and microscopic simulations, Physical Review E, 2000, 62 (2): 1805–1824.
17. P. G. Gipps (1986). A model for the structure of lane-changing decisions. Transportation Research Part B: Methodological, 20, 403–414.
18. R. S. Sutton, D. McAllester, S. Singh, and Y. Mansour. Policy Gradient Methods for Reinforcement Learning with Function Approximation In Advances in neural information processing systems, pp. 1057-1063. 2000.
19. T. Degris, P. M. Pilarski, and R. S. Sutton. Model-free reinforcement learning with continuous action in practice. In American Control Conference (ACC), 2012, pp. 2177-2182, 2012.
20. S. Gu, T. Lillicrap, I. Sutskever, S. Levine. Continuous deep q-learning with model-based acceleration. In International Conference on Machine Learning, pp. 2829-2838. 2016
21. S. Adam, L. Busoniu, R. Babuška. Experience replay for real-time reinforcement learning control. IEEE Transactions on Systems, Vol. 42, No. 2, pp. 201-212.